\documentclass[letterpaper, 10 pt, conference]{ieeeconf}  
\usepackage{stfloats}
\usepackage{graphicx}
\usepackage{caption}
\usepackage{paralist}
\usepackage{color}
\usepackage{amsmath}
\usepackage{amssymb}
\usepackage{multirow}
\usepackage{svg}
\usepackage{svg-extract}
\svgsetup{clean}
\usepackage{hyperref}
\usepackage{booktabs}       
\usepackage{fbox}
\IEEEoverridecommandlockouts                              

\usepackage[subtle, title]{savetrees}
\date{}

\title{\LARGE \bf
Embodied AI with Two Arms: Zero-shot Learning, Safety and Modularity}

\author{
Jake Varley$^{*1}$, Sumeet Singh$^{*1}$, Deepali Jain$^{*1}$, Krzysztof Choromanski$^{*1}$,\\ 
Andy Zeng$^{1}$, Somnath Basu Roy Chowdhury$^{2}$, Avinava Dubey$^{3}$, Vikas Sindhwani$^{1}$ \\
{\small $^{1}$Google Deep Mind Robotics, $^{2}$UNC Chapel Hill,  $^{3}$Google Research *equal contribution}
}

\begin{document}
\twocolumn[{%
\renewcommand\twocolumn[1][]{#1}%
\vspace{1cm}
\maketitle
\vspace{-1cm}
\begin{center}
    \centering
    \captionsetup{type=figure}
    \includegraphics[height=3.5cm]{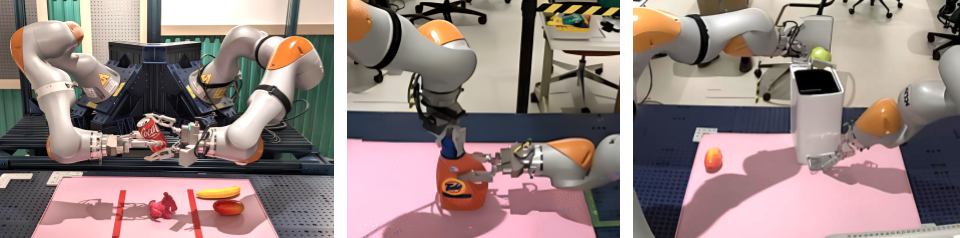}

    \captionof{figure}{\footnotesize \textbf{Left}: Sorting Task; the bi-arm robot follows open-ended long-horizon natural language instructions like {\it ``Move the metal objects to the left side."}. \textbf{Middle}: Bottle Opening, involving a twisting primitive with one arm, and a stable hold of the bottle with the other, where variable compliance is used for the holding and twisting arms. \textbf{Right}: Trash disposal, where the right arm stiffly presses on the foot pedal while the left arm places objects into the bin.}\label{fig:open}
\end{center}%
}]


\thispagestyle{empty}
\pagestyle{empty}
\begin{abstract}
 We present an embodied AI system which receives open-ended natural language instructions from a human, and controls two arms to collaboratively accomplish potentially long-horizon tasks over a large workspace. Our system is modular: it deploys state of the art Large Language Models for task planning, Vision-Language models for semantic perception, and Point Cloud transformers for grasping. With semantic and physical safety in mind, these modules are interfaced with a real-time trajectory optimizer and a compliant tracking controller to enable human-robot proximity. We demonstrate performance for the following tasks: bi-arm sorting, bottle opening, and trash disposal tasks. These are done zero-shot where the models used have not been trained with any real world data from this bi-arm robot, scenes or workspace. Composing both learning- and non-learning-based components in a modular fashion with interpretable inputs and outputs allows the user to easily debug points of failures and fragilities. One may also in-place swap modules to improve the robustness of the overall platform, for instance with imitation-learned policies.
\end{abstract}

\section{Introduction}


The reliability, trustworthiness and safety of emerging AI systems powered by ``foundation models"~\cite{bommasani2021opportunities} is now entering public and policy discourse. Meanwhile, a growing body of work demonstrates how large language models (LLMs) and vision-language models (VLMs)~\cite{ahn2022can,liang2023code,zeng2022socratic,driess2023palm} can endow robots with striking semantic planning and logical reasoning capabilities, enabling them to act on complex instructions received from humans directly in natural language. Such capabilities also raise a new set of challenges: how should semantic and physical safety mechanisms be incorporated all the way from high-level planning to low-level control, so that embodied AI capable of touching and manipulating the physical world can be let loose in human-centric environments? 

For bi-arm robots, in addition to language comprehension and semantic scene understanding, an instruction such as {\it ``put the toys away in the bins"} requires embodied reasoning to decide which arm should pick up an object and whether a handover from one arm to the other is necessary before placement. The physical execution of a plan to accomplish such tasks must satisfy multiple ``classical" safety constraints: avoidance of arm-arm and arm-scene (i.e., objects not currently being manipulated) collisions; joint/velocity/torque limits; moving only within workspace boundaries etc.  Furthermore, household tasks present a multitude of semantic safety constraints, e.g., a knife must never be picked up, a plate must not be placed on a hot stove; a wine glass must only be transferred in upright orientation; a bag that tears easily must be grasped softly; and so on.
\par Our contributions in this paper are as follows. 
We prototype and benchmark a modular bi-arm system, whose schematic is shown in Figure 2. It's four key modules constitute: (i) an instruction-tuned version of a state-of-the-art LLM~\cite{anil2023palm} to map natural language requests to robot code~\cite{liang2023code}, (ii) an open-vocabulary VLM~\cite{minderer2205simple} to parse and segment images and point-clouds from an overhead RGB-D camera to yield the current visual state and segmented object-centric data, (iii) a state-machine based ``zero-shot" \emph{Skills} library encompassing a variety of single- and bi-arm manipulation abilities, and (iv) a \emph{Control} module encompassing a novel SE(3) bi-arm grasping policy, constrained trajectory optimizer for real-time planning across long time horizons~\cite{singh2022optimizing}, and a tuneable, compliant joint-space tracking controller. The bi-arm grasping policy leverages a novel point cloud transformer~\cite{guo2021pct} and is trained entirely in simulation~\cite{erwin2016pybullet} and deployed in the real world to predict cartesian grasp SE(3) poses from point cloud inputs. The trajectory optimizer plans in the combined joint-space of both arms, incorporating various kinematic and semantic constraints, and is also used as a feasibility reasoner to pass textual feedback such as {\it ``cannot reach the object"} back to the human. 

We evaluate our full stack on several manipulation tasks, including bi-arm sorting, bottle-opening, and trash disposal, all of which require coordination between both arms for completing the task successfully. We demonstrate that the inherent modularity of the stack enables (i) non-trivial \emph{zero-shot} performance, (ii) incorporation of different \emph{modalities} of safety, and (iii) interpretability of failures. 

The interaction between the LLM and the lower layers of control may be viewed as an instance of System1-System2 architecture popularized by ~\cite{kahneman2011thinking}. Architecturally, the LLM is a slow  ``thinker" running a multi-billion parameter model off-robot on the cloud and producing logical plans in response to human instructions; while the lower layers of the stack operate in fast control loops executing directly on the robot hardware.  The entire system is \emph{zero-shot} in the sense that it relies on in-context learning capabilities of LLMs and scene understanding capabilities of pretrained VLMs; the lower layers of the control stack are \emph{model-based} -  either trained entirely in simulation or relying on a kinematic model of the robot. This assumption can be relaxed by introducing learnt policies (e.g., via imitation or reinforcement learning) within the \emph{Skills} library.



\begin{figure*}[!t]
\vspace{2mm}
  \includegraphics[width=0.93\textwidth]{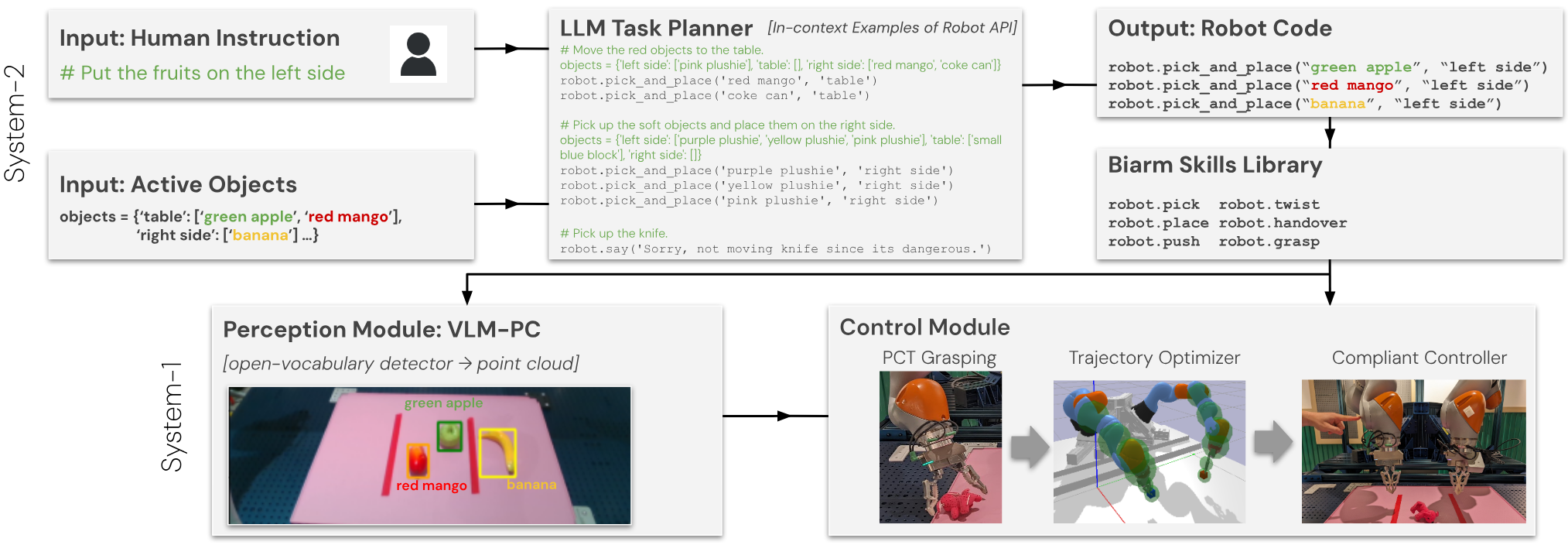}
  \caption{\footnotesize A Modular Bi-arm Embodied AI System. A list of the active scene objects together with the user's instruction, is passed to an LLM Task Planner. Leveraging in-context learning on example API usage, the planner generates a sequence of high-level executable commands for the robot. Each high-level command, e.g., \texttt{pick-and-place}, leverages a state machine that orchestrates VLM-Point Cloud (VLM-PC) perception-based helper functions, and a collection of bi-arm \emph{Skills} such as \texttt{pick, handover, place}. Finally, within each \emph{Skill}, filtered point clouds from the \emph{Perception} module are combined with Point Cloud Transformer (PCT)-based grasping policies and SQP-based motion planning to generate joint-space trajectories for both arms, tracked using a compliant controller.} \label{fig:stack}
  \vspace{-2mm}
\end{figure*}

\section{Related Work}
This work is at the nascent intersection of bi-arm robotics, large foundation models for control, and AI safety. 

{\it Bi-arm Manipulation}: For robots to effectively operate in environments designed by and intended for humans, they arguably need an anthropomorphic form factor. This has led to a surge of recent interest in humanoids and bi-arm systems~\cite{avigal2022speedfolding,hayakawa2021dual,lertkultanon2018certified,zhao2023learning,mirrazavi2018unified}; see~\cite{smith2012dual} for a survey. One prime motivation is to transfer natural bi-arm skills of a human teleoperator to the robot~\cite{edsinger2007two,jau1988anthropomorhic,yoon1999experimental,kron2004bimanual,taylor2010control}. In~\cite{zhao2023learning}  a low-cost puppeteering system is presented for acquiring fine bi-arm skills via a novel imitation learning approach, while~\cite{stepputtis2022system} find that collecting force-torque information in training demonstrations is crucial for achieving high success rates and robustness in tasks such as bi-arm insertion.~\cite{chen2022towards} release a simulation environment covering tens of bi-arm manipulation tasks. In~\cite{avigal2022speedfolding},  bi-arm skills are learnt for folding garments from random initial configurations in under two minutes.

{\it Foundation Models for Robotics}:  With the emergence of common sense reasoning, language understanding and code-writing capabilities, it is no surprise that LLMs were quickly embraced as a primary mechanism for high-level task planning in Robotics. In~\cite{ahn2022can}, an LLM planner is biased to only “say'' what the robot “can” actually do next; this is done by combining LLM likelihoods with affordance values associated with low-level policies (e.g., RT1~\cite{brohan2022rt}) deployed on the robot. This framework can receive further grounding feedback from success detectors~\cite{huang2022inner} and scene descriptions from VLMs~\cite{zeng2022socratic}. Palm-E~\cite{driess2023palm} is a multimodal extension of the Pathways Language Model~\cite{chowdhery2022palm} where image, text and robot state tokens are intermixed in a shared embedding space. Palm-E fine-tunes the base model to generate the next step of the plan indexing a library of language conditioned short-horizon policies~\cite{brohan2022rt}. In~\cite{liang2023code}, it is demonstrated that LLMs are also capable of much more complex orchestration of perception and control APIs by directly generating code snippets, complete with loops, conditionals and subroutine definitions. We remark that while our work also leverages LLMs as a high-level command orchestrator, the contribution is focused on the design of the modular system to address aspects such as motion feasibility and safety, which play a significantly more salient role in the bi-arm setting.

{\it AI Safety}: Meanwhile, it has been noted that such ``robot brains" with capabilities never seen before may also inadvertently inherit biases and fragility of large models, e.g. see~\cite{hundt2022robots,bommasani2021opportunities,zou2023universal} and references therein on AI safety literature. Modular architectures may facilitate explainability and interpretability~\cite{linardatos2020explainable}, which may mitigate such concerns.



\section{A Modular, zero-shot, safe bi-arm system}
Our system - see Figure 2 -  formally comprises of four hierarchical modules that we describe in this section.

\subsection{Module 1: LLM Task Planner}
At the topmost level of our stack, we leverage an LLM to provide an interface between the (human) user and high-level commands for the robot.
Specifically, we prompt an instruction fine-tuned version of a state-of-the-art LLM~\cite{anil2023palm} to map user instructions to  executable code for the robot, an instance of the Code-as-Policies framework~\cite{liang2023code}. Formally, the planner module \texttt{Plan} is invoked as: 
\[
    {\cal P} = \texttt{Plan}\left( {\cal I}, {\cal S} | {\cal C}\right),
\]
where $\cal P$ denotes the generated plan (executable code), $\cal I$ denotes the current user instruction, $\cal S$ denotes the textual description of the current state, and $\cal C$ denotes a fixed preset context. In our setting, the context $\cal C$ corresponds to $n$ (instruction, code) examples $\{({\cal I}_j, {\cal S}_j, {\cal P}_j)\}_{j=1}^n$ that set the stage for in-context learning, see Figure~\ref{fig:stack}. The state $\cal S$ provides a dictionary of observed objects and their locations in the workspace, as determined using the perception module (described later) or explicitly provided by a human operator.

{\it LLM-Robot Interface}: The executable code provided in context and generated by the LLM invokes commands from a  \emph{high-level API} only, namely \texttt{pick-and-place, unscrew-cap, discard-trash}, whose arguments are limited to (i) object and location names, (ii) arm-id (e.g., ``left-arm, right-arm"), and (iii) a few tuneable values. We refer to this LLM-facing API as the `LLMBot API'. Under the hood, this API orchestrates a state-machine composed of pre-defined bi-arm \emph{Skills}, such as \texttt{pick, place, handover, push}, and helper functions such as \texttt{find-object, get-visual-state, update-obstacles}. These functions are parameterized by the strings and values input to the high-level commands. The \emph{Skills Library} constitutes the second key module of this stack.

Note that neither the individual \emph{Skills} nor the helper utility functions are exposed within the LLMBot API to simplify ``prompt engineering". Future work would entail expanding the context prompt to include for instance, the APIs for various \emph{Skills} and utility functions to allow the LLM to directly generate the state-machine logic, similar to~\cite{liang2023code}.


\subsection{Module 2: Bi-arm Skills Library}
Each \emph{Skill} is a combination of various methods from the \emph{VLM-Point-Cloud (VLM-PC)} perception module and parameterized bi-arm \emph{Control Primitives} to generate a sequence of motion plans and gripper commands for both arms. In our framework, a \emph{Control Primitive} corresponds to a function that combines point cloud grasping policies (Section~\ref{sec:pct}) to generate object grasp poses and a constrained trajectory optimization solver (Section~\ref{sec:trajopt}) to plan the collision-free trajectories for the arms. Examples include methods such as \texttt{grasp-at-pose, move-to-pose, move-along-trajectory}. Importantly, we do not leverage any training data from the robot or environment for the design of these skills; they are a combination of an off-the-shelf vision foundation model, simulation-learned policies, and classical trajectory optimization routines. However, the modularity of our system permits replacing or augmenting this library with learned language-conditioned policies, such as those learned via imitation learning~\cite{zhao2023learning}.

\subsection{Module 3: VLM-PC Perception}
\label{sec:vlm}
The perception module processes incoming RGB-D images from a realsense camera as follows. The RGB image is passed through the OWL-ViT \cite{minderer2205simple} VLM to detect bounding-boxes, provided a text description for the object (or part of an object). The extracted bounding-box is used to segment and filter the depth image and extract the target object's point cloud. An advantage of this processing is that the performance of the perception module is independent from the downstream policies and motion planners. Thus, more capable VLMs and/or more flexible segmentation masks, e.g. \cite{kirillov2023segment}, are readily embeddedable in our stack.

A second advantage of using OWL-ViT as our VLM backend is the use of ``teaching" mode. Image datasets for training foundational VLMs do not typically feature extensive labeling on \emph{parts} of an object, focusing primarily on macro detection instead. However, finer manipulation necessarily requires more precise localization of parts of an object, e.g., \emph{a bottle's handle and lid} or \emph{the push pedal on a trash can}. In these settings, we can manually draw a bounding-box around object parts in a handful of different configurations, and subsequently rely on the VLM's image-based retrieval mode to detect future instances. This capability will be distinctly featured in Sections~\ref{sec:exps_bottle} and~\ref{sec:exps_trash}.

\subsection{Module 4: Control}

\subsubsection{Transformers for Point Cloud Grasping}
\label{sec:pct}
We leverage Point Cloud Transformers (PCTs,  \cite{pct}) to define the backbone of the bi-arm grasping policies, defined by a PCT-encoder operating on the filtered point cloud of the target object, extracted from the \emph{VLM-PC} module. The full observation space for the PCT-policy consists of (i) the variable-size $N \times 3$ object PC which is mean-shifted to the workspace origin, (ii) the mean of the pre-shifted filtered point cloud in the workspace frame, and (iii) the major axis of the object points in its frame.

\paragraph*{Action space} The PCT-policy generates an SE(3) grasp pose defined by the fingertip position, relative to the object center, the grasp direction vector in the workspace frame, and a rotation angle about this direction vector. The trajectory for the arm to execute the grasp is generated using the constrained trajectory optimization solver.

\paragraph*{Architecture} A PCT-encoder consists of (i) two initial convolutional layers intertwined with batch normalization layers, (ii) two attention layers with the standard softmax-kernel, and (iii) two convolutional-batchnorm-ReLU blocks (with $\textrm{Leaky-}\textrm{ReLU}$ nonlinearity). Importantly, the outputs of the attention layers are concatenated before being passed to the next layers. This design choice was found effective in the original paper on PCTs (\cite{pct}) and was further validated in our preliminary experiments. One of the tokens is the \textit{class token}; its final fixed-size embedding $\mathbf{e}$ is used as a latent representation for the entire input PC.

\paragraph*{Fusing non-PCT observations}
The other observations, i.e., object center and major axis, are concatenated with the point cloud embedding $\mathbf{e}$ and passed through two dense layers with width $8$ and $\mathrm{tanh}$ activation. Their output is processed by another dense layer to produce the $7$-dimensional normalized action in the range $[-1, 1]$. The vector is de-normalized and projected onto the SE(3) action space.

\paragraph*{Training \& deployment} The PCT-policy is learned in a simulated environment (developed using PyBullet~\cite{erwin2016pybullet}), featuring the bi-arm robot and the table workspace, and Blackbox optimization (\cite{salimans}, \cite{isim2real}). We chose the BGS variant (\cite{isim2real}) with $l=50$ perturbation-directions, ES-smoothing parameter $\sigma=0.02$, step size $\eta=0.02$ and $\tau=30\%$ top directions. In training, the robot only sees $k=5$ different objects: a coke can, water bottle, chalkboard eraser, banana, and an octopus soft toy. The policy is given a binary reward for each object: $1$ if the object position is above the table and close to the gripper after the grasp and $0$ otherwise. 
Performance in sim is measured as an average over $100$ episodes and converges to a $66\%$ grasp success for both left and right arm.
Importantly, we observe no sim-to-real gap in performance and the policies learned in simulation are deployed on the robot with \emph{no additional fine-tuning.}


\subsubsection{SQP Motion Planner}
\label{sec:trajopt}

The second sub-module within a \emph{Control Primitive} corresponds to constrained bi-arm trajectory optimization, solved using the Sequential Quadratic Programming (SQP) solver designed in~\cite{singh2022optimizing}. In particular, we heavily draw upon its adaptation for whole-body planning in~\cite{lew2023robotic} by leveraging bounding-boxes as obstacles, modeling the robot geometry with a set of spheres, and single-integrator joint kinematics for each arm. Together, the set of all constraints constitute: (i) joint position and velocity limits, (ii) workspace limits, and (iii) arm-to-arm, and (iv) arm-to-object collision avoidance\footnote{Note that in context of \texttt{pick} or \texttt{push} \emph{Skills}, the target object is excluded from the obstacle list.}. The cost function is a combination of penalties on goal-reaching for the end-effectors, where the goals are generated by higher-level modules (e.g., the grasping poses) and control effort.

A na{\" i}ve application of constrained trajectory optimization would entail using a long fixed-horizon problem in order to cover a wide temporal range of desired motions. However, given the non-convexity of the problem, this would be prohibitively slow for a real-time system. Instead, we adopt a \emph{quasi-model-predictive-control} (quasi-MPC) approach whereby we solve a \emph{sequence} of fixed but short-horizon problems together, by concatenating the solutions, i.e., the terminal state of the solution of one ``meta-step" becomes the initial state for the next step's problem. To encourage convergence to the desired SE(3) targets for the end-effectors, we apply a telescoping terminal constraint for each meta-step:
\begin{align*}
    e_p^{(i)}(T) \leq \max\{e_p^{(i-1)}(T), \varepsilon_p\},\ \  e_r^{(i)}(T) \leq \max\{e_r^{(i-1)}(T), \varepsilon_r\}
\end{align*}
where $e_p^{(i)}(T)$ and $e_r^{(i)}(T)$ are the terminal position (p) and rotational (r) errors for meta-step $i$, and $\varepsilon_p, \varepsilon_r$ are fixed convergence tolerances. Notice, thus, that \emph{recursive feasibility} of the sequence of problems is guaranteed with this telescoping constraint. The loss of global optimality (as with any receding-horizon approach) is more than compensated for by the drastically improved speed, enabling real-time execution (see Section~\ref{sec:exps_timing} for detailed latency benchmarking).

In practice, we cap the number of meta-steps at 10, and use a prediction horizon ($T$) of 10 timesteps with 0.1s resolution within each meta-step's optimization problem, giving a total temporal range between 1 to 10s.

\subsubsection{Compliant Control}
\label{sec:compliant}
The final sub-module and the lowest-level of the stack is defined by a joint-space compliant torque controller, consisting of a feedforward and PD-feedback law. The feedforward torques are generated by smoothing the desired motion generated by the SQP planner to yield desired joint accelerations and leveraging inverse dynamics. Importantly, the stiffness and damping parameters are tunable and may be used to adjust the compliance of the arms, depending on the task. The controller releases force accumulations when a closed kinematic-chain is formed, for example, during during an arm-to-arm handover, and enables safety during unintended contact with the workspace, or force interactions with a human operator. The computed torques are executed on the two arms in a shared high-rate control loop.


\section{System Experiments}
\label{sec:demos}


We demonstrate our modular stack on a series of tasks with graduated complexity.
For each task, we detail: (i) semantic reasoning and LLMBot API commands, (ii) a sketch of the state-machine and relevant \emph{Skills}, (iii) experiment setup and assumptions, and (iv) results and limitations. Crucially, our stack enables us to \emph{precisely} isolate the cause of each type of failure -- thereby allowing a practitioner to modularize debugging and improvement.

\subsection{Bi-arm Semantic Sorting Task}
\label{sec:exps_sorting}

\begin{figure}[ht]
\centering
      \vspace{3mm}
     \includegraphics[width=0.45\textwidth]{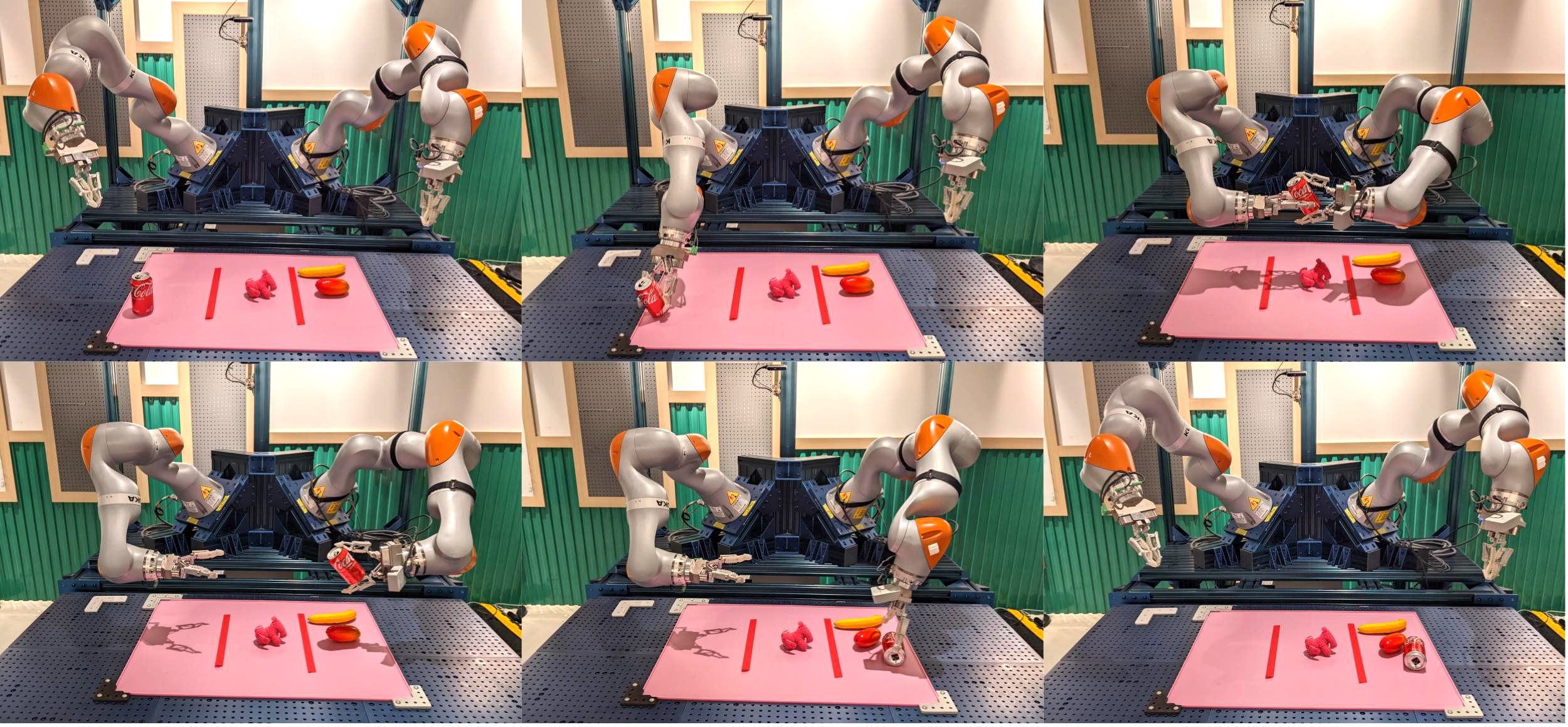}
      \caption{\footnotesize Example execution of a successfully generated bi-arm sorting LLM plan to \emph{``move the metallic objects to the left side"}. The robot switches to picking with the right arm after it fails to reach the can with its left arm, and then performs a \texttt{handover} to the left arm for placing.}
       \label{fig:sorting}
\end{figure}

The objective of this task is to sort $15$ tabletop objects (e.g.,~\textit{banana, apple, sprite can, coke can, small yellow block, purple plushie} etc) with diverse geometry based on a randomly chosen grouping attribute (e.g.,~\textit{cans, blocks, soft objects, metal objects} etc). Target objects are randomly placed in the workspace; the target destination (e.g., ``left side") is a pre-defined area on the table; and the scene is populated with random distractor objects. The transfer from one side to another may  require bi-arm coordination, e.g. with one arm picking and handing over to the other for placement. Specifically, we demonstrate: (i) semantic parsing of the user natural language instruction into an appropriate LLMBot command, (ii) bi-arm kinematic-feasibility reasoning, and (iii) \emph{guaranteed} safe execution, where safety in this task pertains to satisfaction of kinematic and workspace constraints. See Fig~\ref{fig:sorting} for an example execution.

\paragraph*{Semantic Reasoning}
An example $(\mathcal{S, I, P})$ tuple is shown in Figure~\ref{fig:stack}, where the state $\mathcal{S}$ is auto-generated via a \texttt{get-state} function that queries the VLM based on a pre-defined list of all objects. The only relevant LLMBot API is \texttt{pick-and-place}, which accepts two strings as inputs: the object to be moved and the desired location on the table.

\paragraph*{Relevant Skills and State Machine}
The relevant \emph{Skills} within \texttt{pick-and-place} are $\{$\texttt{pick, place, handover}$\}$. The complexity of the orchestration of these skills stems from the conditional logic associated with handling failed or infeasible grasp and/or place attempts. For instance, in case of failed grasps from the PCT-policy, the state machine leverages a backup grasp policy whereby SQP computes the grasping pose as a byproduct of guiding the end-effector to the mean of the target object's filtered point cloud. Secondary backup involves switching the \texttt{pick} arm. Finally, placing on the left- or right-side of the table constrains the placing arm, thereby necessitating tracking the arm which picks up the object and leveraging a \texttt{handover} maneuver if necessary. 

\paragraph*{Experiment Setup}
We procedurally generated 30 sorting tasks and with an LLM context $\mathcal{C}$ consisting of 2 example tuples of $(\mathcal{S, I, P})$, benchmarked (i) planning success rate, (ii) execution success rate, and (iii) failure types. We used the PaLM-2 (340B) instruction fine-tuned LLM~\cite{anil2023palm}. Additionally, we performed an independent evaluation of the sim-trained PCT grasping policy (see Sec.~\ref{sec:pct}) on the robot. We tested the policy on $108$ different scenes, including objects unseen during training.

\paragraph*{Results and Limitations}
 The grasping policy yielded $70$ successful grasps ($65\%$ accuracy) showing no sim-to-real performance drop.
 Of the 30 generated tasks, the LLM yielded 27 semantically correct plans.  Each of the 27 successful task plans $\mathcal{P}$ was generated verbatim from the LLM was then physically executed on the bi-arm cell, with one example shown in Figure~\ref{fig:sorting}. Table~\ref{tab:sorting_results} illustrates the overall success rate and failure types, as a fraction of the successful plans.
\begin{table}[htb]
\centering
\begin{tabular}{cccc} \toprule
\multirow{2}{*}{Success} &  \multicolumn{3}{c}{Failures} \\
\cmidrule{2-4}
& Perception & Handover & Collision \\
\midrule
21 / 27 & 3 / 27 & 2 / 27 & 1 / 27 \\
\bottomrule
\end{tabular}
\caption{\footnotesize Benchmarking success and failure types on the bi-arm semantic sorting task. The execution success rate (as a fraction of successful plans) is $77.7\%$, while the overall end-to-end success rate is $70\%$, accounting for LLM semantic plan generation errors.}
\label{tab:sorting_results}
\end{table}

 The `Perception' errors corresponded to the VLM either failing to find a certain object or mistaking one object for another (e.g., ``a green block" vs ``a sprite can" that has fallen over). The `Handover' failures corresponded to the object being dropped as it is transferred from one hand to the other. The fragility of this skill stemmed from not using a learned policy for this delicate manuever. Finally, `Collision' failures stemmed from an object within a gripper colliding with another object on the table. This error was not unreasonable since we did not account for the grasped object's geometry during post-grasp motion planning. We remark that the modularity of the system permits an in-place replacement of the VLM with a better semantic detector, e.g.,~\cite{kirillov2023segment}, and learned manipulation policies in-place of the grasping policy and trajectory optimizer.


\paragraph*{Incorporating Semantic Safety}
To test LLM planning under semantic safety constraints, we add a {\it red knife} to the set of objects which triggers the ``metallic", "sharp" and ``red objects" attributes. The only other attribute we randomize over is ``soft objects". Remarkably, simply adding the following single (instruction, code) pair to the LLM context resulted in $90\%$ correct plans over $30$ evaluations.

\begin{small}
\begin{verbatim}
# Pick up the knife.
robot.say('Sorry, not moving knife 
since its dangerous.')
\end{verbatim}
\end{small}

So, for example, when instructed to {\it ``Put the red objects on the right side"} with a {\it coke can, small red block} and {\it red knife} in the workspace, the generated code is:
\begin{small}
\begin{verbatim}
robot.pick_and_place('coke can', 'right side')
robot.pick_and_place('small red block',
                     'right side')
robot.say('Sorry, not moving red knife 
since its dangerous.')
\end{verbatim}
\end{small}
With safety prompting, we observed that in response to other instructions, the LLM also starts to generate rationales such as {\it 
``Sorry, not moving sprite can since its not soft"}.
Finally, we observed that $3$-shot safety prompting improves performance over $1$-shot, reaching $93.3\%$ on the procedurally generated tasks.


\subsection{Bottle Opening}
\label{sec:exps_bottle}
We now present a more dexterous task necessitating coordinated bi-arm manipulation: lid-opening. In this task, the robot is presented with a screw-cap container and is required to stabilize the container as it unscrews the cap. The primary features demonstrated in this task include: (i) semantic parsing of \emph{parts} of an object for part-level localization, (ii) coordinated manipulation in a fundamentally unstable setting, and (iii) the role of compliance in the joint tracking controller.

\begin{figure}[h]
    \begin{center}
       \vspace{3mm}
    \includegraphics[width=.45\textwidth]{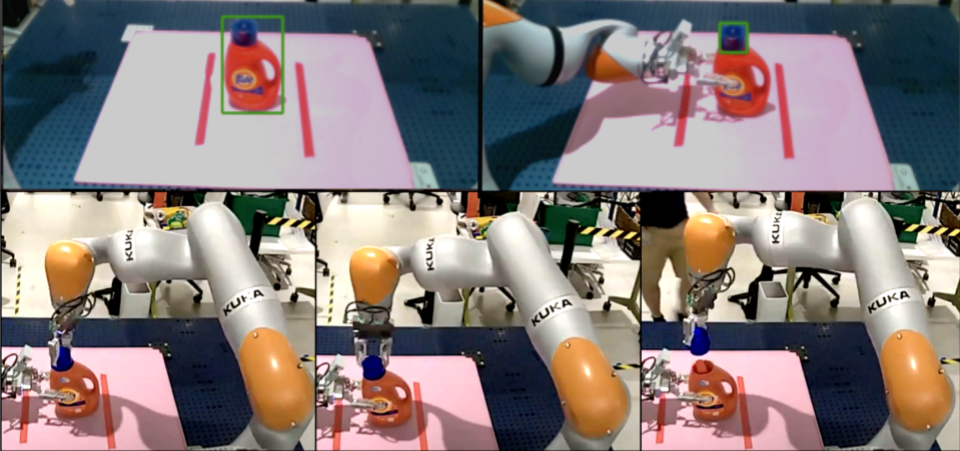}
    \caption{\footnotesize Bottle opening task conducted by our manipulation robot. \textbf{Upper row:} OWL-ViT's part-specific detection of the bottle and cap. \textbf{Lower row:} Execution of the state-machine. Aside from OWL-ViT, no other learning was involved.}
    \label{fig:bottle-opening}
    \end{center}
\end{figure}

\paragraph*{Semantic Reasoning} As with bi-arm sorting, a single command is used within the LLMBot API: \texttt{unscrew-cap}. This command accepts as inputs captions for the container to be opened as well as its cap, and number of desired twists of the cap. The majority of the semantic reasoning is handled by OWL-ViT as we present the robot with a variety of containers, necessitating localization of \emph{parts} of the object using a variety of labels for the container and cap; see Figure~\ref{fig:bottle-opening} for examples of such semantic localization.

\paragraph*{Relevant Skills and State Machine}
The relevant \emph{Skills} needed are simply \texttt{grasp} and \texttt{twist}, where one arm is tasked with grasping and holding the container, while the twisting arm repeatedly grasps and twists the container cap for a preset number of twists\footnote{One may alternatively leverage either vision or force-sensing to detect when the cap is loose.}. The key challenges thus stem from (i) the accuracy of \emph{VLM-PC}'s localization of the object parts, and (ii) the long-horizon nature of the task, with several potentially fragile bottlenecks.

\begin{figure}[h]
    \begin{center}
    \includegraphics[width=.95\linewidth]{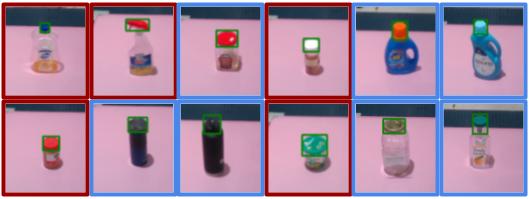}
    \caption{\footnotesize OWL-ViT Lid localizations for held-out bottles, are highlighted with green bounding boxes. Red border indicates bottles the robot failed to open while blue indicates bottles the robot successfully opened.} \label{fig:hold-out-bottles}
    \vspace{-4mm}
    \end{center}
\end{figure}

\paragraph*{Experiment Setup}
This task assumes that (i) the container is always placed upright with the major-minor axes aligned with the workspace's coordinate system, (ii) the left arm will grasp the bottle approaching from the left, and (iii) the right arm will grasp the cap approaching from the top. The set of valid initial conditions could be extended by adding a \texttt{re-orient} skill to the library to first align the container in a pose suitable for further manipulation. We adjusted our state machine for this task using a `Tide' detergent bottle and then \emph{zero-shot} test on 12 \emph{unseen} containers, shown in Figure~\ref{fig:hold-out-bottles}.

\paragraph*{Results and Limitations}
We highlight that 7 / 12 (success rate: $58.3\%$) containers were successfully opened \emph{zero-shot}; see Figure~\ref{fig:bottle-opening} for an example execution. Three failures corresponded to the smallest containers tested - peanut butter jar, paprika jar, and a snack box. For these cases, the proximity of the holding and twisting grippers either resulted in partial opening or bumping the object out of grasp. The `Windex' bottle's unusually shaped lid caused a re-grasp failure. Finally, the soap bottle featured the smallest lid tested, causing a partial open and eventual re-grasp failure.

This zero-shot generalization was achieved with a grasp heuristic (grasp at the mean of the object part's filtered point cloud with preset approach directions). A crucial factor in several of the successful cases was the role of the compliant joint-tracking controller, specifically when the cap is (re-)grasped. Instead of the bottle stabilizing arm remaining rigid, the arm adjusts according to the pressure applied by the cap-twisting arm, allowing a stable kinematic loop to be established between the arms, container, and cap. Notably, this result suggests that further improvements that leverage learning-based policies may benefit from predicting in both cartesian- and joint-space, and feature compliant closed-loop structures.

\subsection{Bi-arm Trash Disposal Task}
\label{sec:exps_trash}
Our final studied task also involves bi-arm coordination: trash disposal. Given a push pedal-type trash can and a list of objects pre-identified to be ``discarded," the robot is required to dispose off the items into the trash can. In addition to the need for coordinated motion, this task also highlighted the need for \emph{part-}level localization, namely, the parts of the trash can.

\paragraph*{Semantic Reasoning} As with the other two tasks, we use a single command within the LLMBot API: \texttt{discard-trash}, which accepts as inputs: the arm-ids for the pushing and placing arms\footnote{This can readily be replaced with similar logic as used within \texttt{pick-and-place} in which the state machine automatically decides upon the picking and placing arms based on the kinematic feasibility feedback from the trajectory optimizer.}, captions for the trash can's lid and push pedal, and a list of objects to be discarded.

\paragraph*{Relevant Skills and State Machine}
This task required three \emph{Skills}: \texttt{pick, place, push}, where the pushing-arm is tasked with keeping the pedal pressed while the placing arm picks and places the objects into the trash can.

\begin{figure}[h]
    \begin{center}
    \vspace{2mm}
    \includegraphics[width=.975\linewidth]{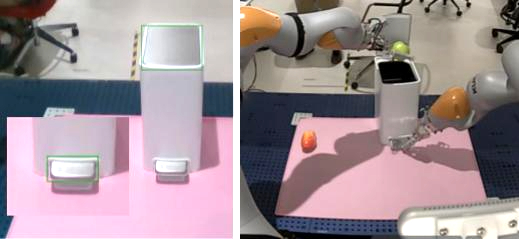}
    \caption{\footnotesize \textbf{Left:} User annotated images to teach OWL-ViT the trash can lid and pedal. \textbf{Right:} Leveraging initial OWL-ViT detections, the left arm pushes on pedal with stiff compliance gains while the right arm places the picked-object at the trash can lid's location. The roles of the arms are \emph{not} fixed, and can be dynamically adjusted by the state-machine based on kinematic feasibility.} \label{fig:trash-disposal}
    \end{center}
\end{figure}

The key challenge of this task stemmed from the localization of the trash can's components: namely the opening and pedal. These parts were unknown to OWL-ViT, with resulting bounding boxes either corresponding to the entire bin or spurious features in the captured image. As a result, we were first required to \emph{teach} the VLM the components of the trash-can; cf. Figure~\ref{fig:trash-disposal}.

\paragraph*{Experiment Setup} Following OWL-ViT teaching, we tested 10 scenarios, each with 1-2 ``garbage" items and 1-2 distractor objects, along with minor randomization of the trash can's location. We fixed the pushing arm as the right arm and the placing arm as the left arm, however permit the pushing arm to assist picking up the trash and performing a handover to the placing arm.

\paragraph*{Results and Limitations} Across the 10 scenarios encompassing 16 total ``garbage" items, 3 items were subject to `Perception' errors after mis-classification by OWL-ViT. Of the remaining 13 items, three bounced off the rim of the can (two of these were due to the banana, which was an inherently difficult item to place correctly), and in one case, the pushing arm's gripper slipped off the pedal. Thus, a total of 9 items were successfully disposed out of the 16; a ``success" rate of $56.25\%$.

The experiment demonstrated the need for compliant tracking controllers and accounting for object geometry post-grasp (e.g., placing the banana correctly in the can). Indeed, despite the target of the pushing arm being below the surface of the push-pedal, a minimum stiffness for the arm's controller was necessary to ensure the pedal would be sufficiently pressed down; notwithstanding the gripper slipping.

\subsection{Latency}
\label{sec:exps_timing}
System Latencies and runtimes are described in Table \ref{tab:latency}.

\begin{table}[htb]
\centering
\begin{tabular}{lllll}
\toprule
LLM   & VLM-PC & PCT   & SQP  & Motion \\
\midrule
0.931 & 0.232  & 0.817 & 1.81 & 20.13 \\
\bottomrule
\end{tabular}
\caption{\footnotesize System Run Times (s). For each module, 30 runs were timed. LLM: sorting task plans. Others: 30 PCT grasps for a scene with a single `pink plushie' toy. SQP timing includes three concatenated motion plans: two for approaching the grasp pose and one for the post-grasp lift, for a total average \emph{predicted plan duration} of 10.7s (i.e., 107 timesteps for the overall trajectory optimization problem, encompassing over 16K inequality constraints).}
\label{tab:latency}
\end{table}

\section{Conclusion}
\label{sec:conclusion}
In summary, this paper presents a modular AI system capable of performing bi-arm tasks based on natural language instructions. The system combines state-of-the-art LLMs for task planning, Vision-Language Models for perception, Point Cloud Transformers for grasping, and constrained trajectory optimization for motion planning. It prioritizes safety considerations by incorporating constraints within motion planning and compliant tracking controllers for human-robot proximity. The system's zero-shot capabilities enable it to perform tasks without specific training on the robot or workspace, demonstrating its potential utility in real-world scenarios with complex instructions and safety constraints. This work paves the way for incorporating additional avenues of interaction with LLMs and VLMs, where user kinesthetic or descriptive feedback may be used in a persistent self-improvement loop via exploiting in-context learning.
This system's modular structure readily enables in-place replacement of any module with newer and better iterations as they appear (ex grasping policies or learned skill primitives).

\subsection*{Acknowledgements}
\label{sec:acknowledgements}
We thank Stefan Schaal and Keegan Go for help on controllers and infrastructure; Michael Ahn, Ken Oslund, and Grace Vesom for infrastructure, and Chad Boodoo for hardware support.
\bibliographystyle{plain}
\bibliography{safe_bimanual.bib}
\end{document}